# Negation Handling in Machine Learning-Based Sentiment Classification for Colloquial Arabic

Omar Al-Harbi
*Jazan University, Saudi Arabia*

**ABSTRACT**

*One crucial aspect of sentiment analysis is negation handling, where the occurrence of negation can flip the sentiment of a sentence and negatively affects the machine learning-based sentiment classification. The role of negation in Arabic sentiment analysis has been explored only to a limited extent, especially for colloquial Arabic. In this paper, the author addresses the negation problem of machine learning-based sentiment classification for a colloquial Arabic language. To this end, we propose a simple rule-based algorithm for handling the problem; the rules were crafted based on observing many cases of negation. Additionally, simple linguistic knowledge and sentiment lexicon are used for this purpose. The author also examines the impact of the proposed algorithm on the performance of different machine learning algorithms. The results given by the proposed algorithm are compared with three baseline models. The experimental results show that there is a positive impact on the classifiers' accuracy, precision and recall when the proposed algorithm is used compared to the baselines.*

## INTRODUCTION

Sentiment analysis or opinion mining is the process of automatically identifying the opinions expressed at the level of a word, sentence, or document. In recent years, Sentiment Analysis has received much attention from researchers, and considerable progress has been achieved for different languages, especially for English. However, as this work concerns with the Arabic language, the task of sentiment analysis is still limited. The challenges that face Arabic sentiment analysis are related to the inflectional nature of the language itself. El-Beltagy and Ali (2013) highlighted many issues of sentiment analysis in the Arabic language such as the presence of dialects, lack of Arabic dialects resources and tools, limitation of Arabic sentiment lexicons, using compound phrases and idioms, etc. In general, the Modern Standard Arabic (MSA) and colloquial Arabic are the most commonly used forms of Arabic language in social networks, blog, and forums.

One of the several challenges that face sentiment analysis, in general, is the negation. In sentiment analysis, negation words can reverse the meaning of a sentence; as a result of which the sentiment orientation would be changed. For example, negation words (such as "not") flip the sentimental orientation of the terms (such as "good"). According to Wiegand, Balahur, Roth, Klakow, and Montoyo (2010), negation is a complicated issue, and it is highly relevant for sentiment analysis. This is a complicated issue because it requires detecting the negation words and then identifying the affected words (which are called negation scope) based on either syntactic or semantic representations (Ballesteros et al., 2012). For sentiment classification, there are two major approaches used in the literature. The semantic orientation approach in which sentiment lexicons and other linguistics resources are used to compute the sentiment polarity of a given sentence based on the polarity of its words (Taboada, Brooke, Tofiloski, Voll, & Stede, 2011). The other approach is a machine learning-based sentiment classification, which uses annotated data from which a set of features is extracted as a training data used by a classifier to build a model for predicting the classes of a testing data using one of the machine learning algorithms (Pang, Lee, & Vaithyanathan, 2002). In the literature of sentiment analysis, machine learning approach has outperformed the semantic orientation approach in several aspects (Morsy, 2011). However, the performance of machine learning algorithms would be affected by the presence of negation terms (Jia, Yu, & Meng, 2009; Wiegand, Balahur, Roth, Klakow, & Montoyo, 2010; Zhang, Ferrari, & Enjalbert, 2012). In fact, determining the sentiment polarity requires taking linguistic context (such as negation) into account instead only simple presentations such as Bag-of-Words (BOW) and n-grams. Many studies have been published to address the problem of detecting negation words and the negation scope to improve the performance of machine learning algorithm-based sentiment classification such as in (Jia, Yu, & Meng, 2009; Morante & Daelemans, 2009; Pang & Lee, 2004; Polanyi & Zaenen, 2006).

Likewise, negation plays a crucial role in performance of sentiment analysis for the Arabic language, whether in MSA or dialects (Duwairi & Alshboul, 2015; Morsy, 2011). However, the problem of negation algorithms has been less explored in Arabic sentiment analysis using machine learning. Most of the previous work on Arabic sentiment classification used various machine learning algorithms without considering the effect of negation on the classification performance. Not taking negation into consideration would create a similar representation of two sentences like "أنا احب هذا الفلم" (I like this movie) and "أنا لا احب هذا الفلم" (I do not like this movie), although the first one hold a positive sentiment polarity while the second one holds a negative sentiment polarity. That would negatively affect the performance of the classifiers used for sentiment analysis.

In Arabic sentiment analysis, colloquial texts were less addressed compared to MSA. This is presumably due to the availability of different linguistic resources for the MSA language compared to colloquial Arabic language. However, when it comes to social networks or reviewing products, most people use their dialects instead of MSA. Therefore, in this research, we are concerned with handling the negation problem in the colloquial Arabic texts. Unlike MSA, the colloquial Arabic is not restricted with grammatical rules through which the negation can be simply detected. The poor grammatical structure and the lack of resources tools such as morphological parser, part of speech (POS) taggers, and stemmers for colloquial Arabic made the task of negation handling is challenging.

Previous studies on Arabic sentiment analysis- as will be explained in section 2- adhere to more or less similar approaches to handle the negation problem. These approaches dealt with the problem in different ways; a statistical way in which the frequency of the negation terms in a given sentence is represented as a feature, or another way by negating a window of words whenever preceded by a negation word, by marking them with a negation tag as suggested by Das and Chen (2001). By using these approaches, negation cannot be properly modeled, that can be explained by the following example, a sentence like "التعامل من أرقى ما يكون" (dealing with customers one of the finest) would result with 1 occurrence of "ما" based on the former approach, despite the term "ما" in this example is a relative pronoun not a negation word. Based on the latter approach, in a sentence like "ما في ازعاج بالعكس هادئة جدا" (there is no noise, on the contrary, it was very quiet), the words after "ما" would be marked with a negation, although the only word which supposed to be affected is (ازعاج) "noise". This would result with new features created that negatively affect the performance of the sentiment classification. Nevertheless, we present these models as baselines to be compared with the proposed algorithm. On the contrary, the proposed algorithm aims to detect only the affected words as some opinionated words might not be affected even though they are within the scope of a negation term. Another approach used to capture negation is using higher-order n-grams, such as using bi-gram in the work of Pang, Lee, and Vaithyanathan (2002). Although this approach is convenient, this would fail in cases in which the affected words are at a distance from the negation words. For instance, in a sentence like " لا يوجد بهذا المطعم اي شيء زاكي" (there is no anything in this restaurant is delicious), the algorithm needs 6-gram to capture negation (لا ... زاكي) "no…delicious", and using such high order n-grams would lead to very sparse representation that makes the learning from training data is harder.

As a main contribution, we propose an algorithm that can detect and handle the negation problem in the colloquial Arabic reviews to improve the performance of the machine learning-based sentiment classification. The author also examines the effect of the proposed algorithm on four of the most common classifiers used in sentiment analysis; they are Support Vector Machine (SVM), Naïve Bayes (NB), k-nearest neighbor (KNN), and Logistic Regression. Additionally, a comparison is carried out between the classifiers when our algorithm is used and three baseline models that differ in their methods of determining the negation scope. The proposed algorithm uses crafted rules, linguistic knowledge, and sentiment lexicon. The rules were crafted based on observing many cases of negation in colloquial reviews. It detects the negation words, like (لا، مو، مش) "no, not, not", and then mark the opinionated words that might be affected within a predefined window length of words. These rules do not rely on grammatical knowledge about the relationships between different constituents, as there are no standard grammatical rules to dialectal texts. A major challenge in this respect is determining the sequence of words in the sentence that might be affected (negation scope) by a negation term. Unlike the Arabic language, several approaches based on various aspects of contents have been presented to address this issue in the English language. These approaches require an annotated negation dataset which is not available for colloquial Arabic language. This issue is beyond the scope of this paper; therefore, we solely use a predefined window length of five words that directly follow a negation word.

The remainder of the paper is organized as follows. Section 2 presents related work that considered negation in Arabic sentiment analysis. In section 3, we introduce the methodology through which the sentiment classification including negation handling is performed. We discuss experimentations and results in section 4. Finally, Section 5 presents the conclusion of this work.

## RELATED WORK

Many works have explored the negation problem in detail in the English and other languages (Amalia, Bijaksana, & Darmantoro, 2018; Gautam, Maharjan, Banjade, Tamang, & Rus, 2018; Mittal, Agarwal, Chouhan, Bania, & Pareek, 2013; Villalba-Osornio, Pérez-Celis, Villasenor-Pineda, & Montes-y-Gómez; Zou, Zhu, & Zhou, 2015), whereas few studies have addressed this issue in the Arabic language as this field is still at an early stage. In this section, we explore how the negation problem has been addressed in previous studies of Arabic sentiment analysis. Several studies in sentiment classification have employed different machine learning algorithms. Unfortunately, a few only have considered the negation problem when the terms are turned into features for the learning process. One of the common methods to include the negation into the features representation is to identify all the words whenever they are preceded by a negation word within either a fixed window size, until the first punctuation, or until the end of a sentence, and then have them marked a negation tag, such as in the studies of (Abdulla, Ahmed, Shehab, & Al-Ayyoub, 2013; Adouane & Johansson, 2016; Al-Obaidi & Samawi, 2016; Duwairi, Marji, Sha'ban, & Rushaidat, 2014; El-Halees, 2011). However, none of these studies reports how the sentiment classification was affected by negation or sufficient details about how they handled the problem. According to Wiegand, Balahur, Roth, Klakow, and Montoyo (2010), this method cannot properly model the negation scope, as that would lead to tag words which are not supposed to be. Consequently, one word would be treated as two different words, then feature space increases.

Another method to deal with negation in the Arabic sentiment classification is the frequency or presence of negation words as a feature of the phrase in a supervised classifier such as in the studies of (Adouane & Johansson, 2016; Al-Harbi, 2017; Farra, Challita, Assi, & Hajj, 2010; Hamouda & El-taher, 2013). In this method, the count of sentiment words also can be used as features after reducing the negated words form a polarity type and consider it with the opposite polarity type. One way to obtain the knowledge about sentiment words is by using a sentiment lexicon which contains a list of opinionated words attached with polarity labels. However, this method does not consider how and what the words in the negation scope that should be affected by the negation terms and that would lead to lower performance.

Other few studies have investigated negation in Arabic sentiment analysis. For example, in the work of Elhawary and Elfeky (2010), they propose a technique that handles negation embedded in Arabic reviews. The authors do not mention whether the reviews written with MSA or dialect. They assume that the negation scope is all words whenever preceded with negation terms until the end of the sentence. As explained before, such a method will lead to issues like negating words that should not be affected by the negation terms. Furthermore, no details provided about the negation words they used except its number, which was 20 words, nor how they dealt with the cases in which the negation words do not have the negation sense. As well as, the effect of negation on sentiment classification is not reported. Another work also introduced by Mostafa (2017), which focus on handling negation in both MSA and dialectal Arabic. The author did not mention details about the dataset and dialect used in this work. He provided details about an algorithm that deals with the negation problem; however, what proposed is not different from the traditional methods that inverse any polarity expression that follow a negation term. The author also considered only one case in which exceptional negation is used. The reported results show an improvement after applying the exceptional negation algorithm to the classifiers SVM, NB, and K-NN. On the contrast, our algorithm uses sentiment lexicon to detect only the affected polarity terms within a fixed window size, and many cases appeared in the texts were investigated to be handled.

The work of Duwairi and Alshboul (2015) also focused on the negation problem. In this work, they introduce an unsupervised sentiment analysis for MSA language that includes a morphological framework for negation. They propose a treatment of negation by using a set of rules derived from formal linguistic knowledge. The negation words are categorized into two groups, the first one include (ما، لا، لم، لن) which affect only the verb that appears immediately after them, and the second group contains (ليس) which affect only the two nouns following it. ArSenl lexicon and an Arabic morphological analyzer were used to assign the sentimental value and POS for the terms, respectively. Unfortunately, these rules cannot be applied to the texts in our work due to the presence of dialect that does not abide by the same linguistic rules. Furthermore, the authors did not provide any details about the experiment or the evaluation results for their approach.

## METHODOLOGY

The aim of this research is to handle the negation problem to improve the sentiment classification for the colloquial Arabic reviews. This section describes the methodology through which the proposed algorithm was developed. The following sections introduce the proposed algorithm, necessary resources, and tools.

### Dataset

To train the classifier, we need an annotated dataset. For the purpose of this work, the author used a publicly available dataset for Jordanian dialect presented by Al-Harbi (2017). The dataset is annotated on the document level, and it considers only two polarity classes, which are positive and negative. To balance the dataset, we randomly selected 2400 reviews, of which 1200 were positive, and 1200 were negative. The data consists of MSA and colloquial Jordanian reviews about various domains (restaurants, shopping, fashion, education, entertainment, hotels, motors, and tourism).

### Pre-processing

The pre-processing stage included removing noise from data, normalization, and tokenization. The process of removing noise from data includes removing misspellings, repeated letters, diacritics, punctuations, numerals, English words, and elongation. After that, a normalization process was applied to particular letters, for example the letters (آ, إ, أ) were converted to (ا), the letters (ى, ئ) were converted to (ي), the letter (ة) was converted to (ه), and finally the letter (ؤ) was converted to (و). Tokenization is the process of dividing a given text into a set of words (tokens) which are separated by spaces.

## Sentiment Lexicon

In this work, the author adopted the dialectical lexicon developed by Al-Harbi (2017). This lexicon consists of 3400 sentiment-bearing words labeled with either positive or negative polarities. The terms were extracted from a text written in both MSA and colloquial language. This lexicon is used to trace the words affected by negation words within the negation scope. During the analysis process, the algorithm will detect all the polarity words and decide whether the negation is meant to positive or negative words, and based on that, only affected words will be reversed.

## Negation Terms List

The author manually collected the most common negation terms used in the reviews and stored them in a list, including different morphological forms of some words. The negation list contains 50 terms, including the terms used in both types of Arabic, MSA such as (لم، ليس) and the dialectal words. In Jordanian dialect, negation is expressed with different terms from MSA. For example, the terms (مو، مش، مهوش، فش، مفيش، مهو) were used in the collected texts. Another way used to negate words is using terms like (مافي، مابيها، مافيها، مافيو، لايمت) due to that the people tend to not space between the negation terms and the following word. We treated such cases as one expression that belongs to the negation words. In this work, if a negation term is detected in the review, the following words within a window length of 5 words will be checked against the sentiment lexicon to decide if they need to be reversed by marking them as negated words. On the other hand, there are several cases in which the negation terms were detected, but not followed by sentimental words, for instance, the review (مافي قسم الأدوات المنزلية) "there is no home appliances section", in these cases, the algorithm will not mark any word within the scope with a negation tag.

## Negation Handling

The main objective of the paper is to address negation in colloquial Arabic reviews to improve sentiment classification. This section describes the proposed algorithm to handle this problem. The algorithm was developed using Python 3.0 programing language, see Figure 1. The input to our algorithm is a review with one or more occurrences of negation terms and output the review with negated polarity words if detected within the negation scope. First of all, we introduce the mechanism of detecting the negation terms and negation scope, which is simply tracing the negation terms within a given review based on the predefined negation terms. Then, if sentimental words are detected within the negation scope, the words will be marked with a negation tag, for instance, (لا <أحب_!> هذا المطعم) "I don't like_! this restaurant". Each negation term is assumed to have a scope of negation effect. In this work, the negation scope is the five words that directly follow the negation term. Determining the negation terms is not an easy task, particularly in the Arabic language since sometimes a negation term in a review does not have the negation sense, or might affect one sentimental polarity without the other. Knowing that, there is no morpho-syntactic tools can be used to the colloquial Arabic, made detecting such exceptions even complicated task. To this end, many cases have been analyzed to come up with rules that can detect such exceptions. In this section, we summarize several cases of how negation terms used in the colloquial Arabic reviews, from which we crafted the required rules to detect negation properly.

**Case1:** a sentence has a negation word followed by an exceptional word (إلا) "but, or except" and polarity expression within the negation scope, and the index of the exceptional word is greater than the index of negation word and less than the index of polarity term like in the sentence (بصراحه ما لقيت الا المعامله الكويسه والاحتراف) "Frankly, we did not find anything but proper treatment and professionalism". In this case, the negation word is used to emphasize whatever the polarity comes after the exceptional word which is positive polarity in this sentence expressed by (الكويسه, الاحتراف) "proper, professionalism". Therefore, the algorithm will not mark the polarity words as negated.

**Case2**: Another phenomenon used commonly in the texts is the use of superlative and comparative words preceded with negation words to express the sentiment as in the sentence (تجربه مافي احلى منها نظيف) (واستقبال جيد) "There is no more beautiful than this experience; it was a clean and good reception". The negation word (مافي) "there is no" followed by the word (احلى) "more beautiful" were used to express positive sentiment, so expectedly any sentimental term comes after those agree with the same polarity and that obvious with the words (نظيف، جيد) "clean, good" that also express positive sentiment. Another example with a negative sentiment (مافي اسوأ من هيك ناس كذابين) "there is no worse than such people, liars", where is the polarity of the word (كذابين) "liars" agrees with the polarity of the word (أسوأ) "worse" and the negation here would not be appropriate. As can be noted in this case, the index of superlative and comparative words is always greater than the index of negation word and less than the index of polarity term. In this case, the algorithm will discard negating the polarity word, and in order to do that, given that we decided to not use any morphological analyzer, we collected and stored the most common used comparative and superlative words such as (أروع، أحلى، أفضل، أجمل، أحسن، أفخم، أرقى، أسوأ، ألعن).

**Case3**: a sentence has two or more sentimental words with different polarities (positive and negative), which fall into the negation scope like in the sentence (مش حلو المكان وسخ بالمرة) "Not a lovely place, it is very filthy". The presence of a negation term in a sentence does not mean that all its polarity words should be affected. As we can see in the example, there are two sentimental words within the negation scope (حلو) "lovely" which expresses positive sentiment and (وسخ) "filthy" which expresses negative sentiment. In this case, the algorithm will detect the polarity of the first sentimental word occurs after the negation term which is in the above sentence (حلو), then will negate only the words that fall into the same polarity within the scope and discarding any other polarity.

**Case4**: a sentence has the negation term (ما) that holds different senses other than the negation, such as interrogative or relative pronoun. For instance, (كل ما نروح عليهم نتنكد ونغير المكان) "Every time we visit them, we got miserable, and we then change the place", based on the discourse context, the word (ما) is a relative pronoun that does not has a negation effect on the negative sentiment of the word (نتنكد) "got miserable"; however, the capability to recognize such cases is hard without a morpho-syntactic analyzer. As mentioned before, we cannot use such analyzer since the available ones have been trained only on MSA. Therefore, we collected and stored all the words that used frequently before or after (ما) when it does not expresses the negation sense. Table 1 shows most the cases of (ما) as not a negation term, whenever, these cases detected the algorithm will ignore negating any polarity term within the scope.

*Table 1. Words might appear after or before the negation word "ما"*

| Negation Term | Cases |
| --- | --- |
| Before "ما" | إن ما، زي ما، نوعا ما، بعد ما، قبل ما, كل ما، مثل ما، مثل ما، حسب ما، قد ما، بدل ما، منذ ما، لحد ما، أول ما، شو ما، اي ما، بدون ما |
| After "ما" | ما بعد، ما قبل، ما شاء الله |

**Case5**: a sentence has the negation term (غير) which in some cases does not have the negation effect on the words like in the sentence (أماكن مناسبة للعائلات غير عن الأماكن المزعجة) "These places are suitable for families; they are different from the noisy places". The word (غير) in the sentence means "different from", and it cannot play the role of the negation on the polarity word (المزعجة) "noisy". In this case it is hard to recognize the word without morphological knowledge, however, the proposed algorithm can handle this case based on knowledge of the words used frequently whether before or after (غير). Those words were observed and collected from the dataset to be fed to the algorithm, Table 2 shows the words.

*Table 2. Words might appear after or before the negation word "غير"*

| Negation Term | Cases |
| --- | --- |
| Before "غير" | لا غير |
| After "غير" | غير عن، غير انو، غير أنه، غير ذلك، غير انتم، غير هيك، غير شكل |

**Case6:** other cases were observed in which the negation terms do not have the negation sense. To enable the algorithm to detect such cases, we collected the words that might frequently appear before or after the negation terms in these cases as knowledge to guide the algorithm to decide whether it is a negation word or not. Table 3 shows the cases we collected along with examples.

*Table 3. Words might appear after or before the negation words "مش، مو، لا، لم"*

| Negation Term | Before | After | Example |
|---|---|---|---|
| مش<br>Not | إذا<br>If | - | افضل منظمين الاعراس بالاردن اذا مش احسنهم.<br>Good wedding organizers in Jordan, if not the best. |
| مو<br>Not | - | متل، مثل<br>like | اسعار جدا طبيعيه مو متل باقي المحلات غاليين.<br>Prices are reasonable, not like other expensive stores. |
| لا<br>Not | - | بد<br>Must | دايما فاصل اعلاني يعني لا بد منه الإزعاج.<br>Always ads during the movie, is it must be annoying? |
| لم<br>Not | إن<br>If | - | أحسن المكتبات بعمان إن لم تكن الأفضل بالأردن.<br>The best library in Amman, if not the best in Jordan. |

*Figure 1. Algorithm for negation handling*

```
Algorithm 1: Negation Handling.
   Input: A Review
   Output: A Review With Negated Polarity Terms If
           Detected
1  Read Review
2  for word in range(len(the review)):
3      if the word is a negation term, and not
4  Case1,
5      and not Case2, and not Case3, and not Case4
6      and not Case5, and not Case6:
7          for i in range(0, 5):
8              if the next word in the sentiment
9                lexicon:
10                 word = word+"_!"
11             else:
12                 Discard and check the next word
13         else:
14             Discard and check the next review
15 close()
```

## Features Representation

In any machine learning approach-based classification, a suitable text representation model is required. This model is often called a vector model or feature model, which is represented by a matrix of term weights. The work of Al-Harbi (2019) has already examined the best text representation for this dataset. Furthermore, the effect of stop words removal, weighting schemes, and stemming (light stemming and root stemming) on the performance of the classifiers were evaluated. The result was a combination of uni-grams, Term Frequency-Inverse Document Frequency (TF-IDF), and stop words removal gives the best performance.

## Sentiment Classification

The goal of the classification is to categorize input data into predefined classes and produce a model based on training data, which predicts the target values of the test data. This work concerns only with two classes; they are positive and negative. In this work, four machine learning algorithms which represent diverse approaches were used to explore the effect of negation handling on colloquial Arabic sentiment classification, namely, SVM, NB, KNN, and Logistic Regression.

## EXPERIMENT AND EVALUATION

This section describes the experiments undertaken to evaluate the performance of the chosen machine learning algorithms on colloquial Arabic sentiment classification when the proposed algorithm is used.

## Experiment Settings

Different experiments were performed to examine the effect negation on the colloquial Arabic sentiment analysis. To accomplish these experiments, we used Rapidminer, which is a software platform that includes a valuable set of machine learning algorithms and tools for data and text mining. As mentioned earlier, the dataset includes reviews which were annotated on the document level, and consist of 2400 reviews of which 1200 were positive, and 1200 were negative. Based on an investigation into the dataset to compute the percentage of the negation, it was found that 47% of the reviews contain explicit negation terms of which 74% were negative, and 26% were positive. It is clear that users tend to use negation terms more when they express negative opinions. The experiments were implemented using four classifiers, namely, SVM, NB, Logistic Regression, and K-NN. Regarding the SVM classifier, we used LIBSVM (Chang & Lin, 2001) with a kernel type of linear as it empirically gave the best performance in our previous works (Al-Harbi (2017); Al-Harbi, 2019). Also, the author investigated K-NN algorithm to find the value of K with which it gives the best performance, based on that the value of K was set to 50. Another issue arises when it comes to using machine learning classifiers is tuning the hyperparameters which can lead to different results for the same classifier. However, finding the optimal hyperparameters is not within the scope of this work; therefore, we follow the default settings provided by RapidMiner software. As previously mentioned, TF-IDF weighting scheme is used to represent the uni-grams after removing the stop words from the reviews.

To examine the effect of the proposed algorithm, there is a need to provide a proper baseline to be compared with. The author used the traditional models that have been employed in different related studies as baseline models for this work. In particular, there are three baseline models to be compared with the proposed algorithm. The first one is baseline1 in which the simple uni-gram model is used without considering the negation problem. Secondly, baseline2 in which a uni-gram model is used, considering the negation problem with a negation scope of five words that directly follow a negation term, where, each term within the scope will be tagged with the negation mark. The last one is baseline3, in which a uni-gram model is used with a negation scope includes all the words that follow a negation term until the end of the sentence, where, each term within the scope will be tagged with the negation mark.

## Evaluation Metrics

In order to evaluate the performance, the N-fold cross validation was employed with N=10, since it has been widely used in this field as it is a reliable technique for assessment. By using this assessment method, the whole dataset was divided randomly into 10 sets with equal sized samples, where the classifier was trained on 9 sets and the remaining set was used for testing. To measure the performance of the machine learning classifiers, the following evaluation metrics were chosen: Accuracy, Precision, and Recall; see Equations 1, 2 and 3. The accuracy represents the correctness percentage of the model by averaging the correct classifications on the total number of classifications. The precision calculates the accuracy of the classifier in regards to the specific predicted class. The recall is sometimes shows the percentage of the correct predicted classes among the actual class in the data.

$$\text{Accuracy} = \frac{TP + TN}{TP + FP + Tn + FN} \qquad (1)$$

$$\text{Precision} = \frac{TP}{TP + FP} \qquad (2)$$

$$\text{Recall} = \frac{TP}{TP + FN} \qquad (3)$$

Where TP indicates a true positive which means the number of the inputs in data test that have been classified as positive when they are really belong to the positive class. TN indicates a true negative which means the number of the inputs in data test that have been classified as negative when they are really belong to the negative class. FP indicate a false positive which means the number of the inputs in data test that have been classified as positive when they are really belong to the negative class. FN indicates a false negative which means the number of the inputs in data test that have been classified as negative when they are really belong to the positive class.

## Results

In this section, the author reports the assessment results of the machine learning classifiers that were used to examine the effect of the proposed algorithm. As mentioned above, three baseline models were used for the comparison with the proposed algorithm. Table 4 displays the performance results of the four classifiers when the baseline models the proposed algorithm are used. It can be noticed that all the classifiers give the lowest accuracy, precision and recall when baseline1 is used in comparison with baseline2. That might be fair because in this model, no linguistic knowledge was involved in the learning process. Another notice can be seen; when we compare the recall and precision of the baseline classifiers, it is found that all classifiers give a lower recall percentage. That can be explained by the fact that mentioned earlier, which is the presence of negation in negative reviews more than its presence in positive reviews. In other words, the false negative FN is bigger than the false positive FP, where FN affects the recall metric and FP affects the precision metric. However, the NB classifier was an exception, where it gave a better recall when the baselin1 is used.

In terms of baseline2, SVM, NB, and logistic regression gave better results of accuracy, precision, and recall compared to baseline1. This improvement obtained when negation is considered by marking all the words within the window size of 5 words. In this case, as we can see, there is a considerable improvement in recall which suggests that considering negation positively handled the false negative FN, and that is again because most of negation appears in negative reviews. Conversely, the K-NN classifier is negatively affected by considering negation in baseline2. This can be seen from the dropping of accuracy and recall compared to their values when baseline1 is used. Additionally, despite the improvement obtained by applying baseline2, that would compromise the learning process by adding useless sparse feature space. For instance, 14304 features have been created when baseline2 is used, on the other hand, there were 12074 features used in the learning process when baseline1 is used.

When baseline3 is used, the SVM classifier does not appear to have a significant improvement in terms of accuracy and precision compared to basline1 and baseline2. However, SVM obtained the best recall, which suggests that considering negation would improve the performance but unfortunately with compromising other aspects. Likewise, considering negation has a negative impact on K-NN, where the performance dropped even less than baseline1. Nevertheless, NB, and logistic regression still give better performance than baseline1, but compared to baseline2, it gave lower results. The low performance of baselin3 can be explained by the issue of sparse representation and its effect on the learning process, where the created features were 18692.

On the other hand, the performance of classifiers using the proposed algorithm showed superiority compared to the baseline models. However, there was an exception; the NB classifier gave a lower performance by a slight percentage compared to baseline2, even though it outperformed baseline1 and baseline3. Another notice worth mentioning that is although the proposed algorithm improved the performance of the SVM classifier, it yielded a little improvement in terms of the accuracy compared to the baseline2. Nevertheless, the algorithm appears to have a significant positive effect on both recall and precision in comparison with all the baselines without compromising each other. Apparently, the same scenario of SVM happened to the logistic regression classifier, where there was a significant improvement of the performance in comparison with baseline1 and baseline3, on the other hand, there was a slight improvement compared to baseline2. Also, we can notice that both recall and precision using the algorithm yielded the best results compared to the baselines. The proposed algorithm also succeeded to improve the performance of The K-NN compared to the baselines. Although the positive impact on the recall of K-NN, the recall was lower than the precision in all cases. It appears that the results when baseline2 is applied were close to our algorithm, with superiority to our algorithm in terms of recall and precision in most cases. Additionally, in contrast to baseline2 and baseline3, our algorithm avoided creating a sparse representation, which would negatively affect the learning process.

*Table 4. Results of proposed algorithm and baseline models*

| Classifier | Baseline Model | Accuracy | Precision | Recall |
|---|---|---|---|---|
| SVM | Baseline 1 | 87.83% | 88.35% | 87.17% |
| | Baseline 2 | 89.08% | 88.47% | 90.00% |
| | Baseline 3 | 87.42% | 85.12% | **90.75%** |
| | Proposed | **89.17%** | **89.10%** | 89.33% |
| NB | Baseline 1 | 77.83% | 76.77% | 79.92% |
| | Baseline 2 | **80.62%** | 78.62% | **84.25%** |
| | Baseline 3 | 79.00% | **79.01%** | 79.25% |
| | Proposed | 80.04% | 78.44% | 83.00% |
| Logistic Regression | Baseline 1 | 83.33% | 84.17% | 82.25% |
| | Baseline 2 | 85.29% | 87.00% | 83.08% |
| | Baseline 3 | 83.67% | 85.32% | 81.42% |
| | Proposed | **85.75%** | **87.02%** | **84.17%** |
| K-NN | Baseline 1 | 86.50% | 87.91% | 84.67% |
| | Baseline 2 | 85.88% | **90.29%** | 80.42% |
| | Baseline 3 | 82.75% | 87.26% | 76.75% |
| | Proposed | **87.75%** | 89.97% | **85.00%** |

## CONCLUSION AND FUTURE WORK

Based on the experimental results, we conclude that using the proposed algorithm for negation handling have a positive impact on machine learning-based colloquial Arabic sentiment classification, yet is far from perfect. The experiments were conducted using SVM, NB, K-NN, and logistic regression, which showed a significant improvement in their performance after applying the negation handling algorithm. The proposed algorithm is rule-based, and the rules were crafted based on observing many cases of negation and simple linguistic knowledge. These rules showed the capability of deciding when the negation should be applied even though the absence of morphological knowledge for colloquial Arabic texts.

In future work, we plan to enable the algorithm to deal with implicit negation that also can negatively affect polarity classification. Another problem that needs to be addressed is that the usage of intensifiers and diminishers, which can change the polarity of words or phrases.